\documentclass[10pt,journal,compsoc]{IEEEtran}
\ifCLASSOPTIONcompsoc
  \usepackage[nocompress]{cite}
\else
  \usepackage{cite}
\fi
\ifCLASSINFOpdf
\else
\fi
\usepackage{amsmath}
\usepackage{amssymb}
\usepackage{mathtools} 
\usepackage{xcolor}
\usepackage{comment}
\usepackage{ragged2e}
\usepackage{makecell, caption}

\begin{document}
%
\title{Dynamic multi-object Gaussian process models: A framework for data-driven functional modelling of human joints}
%
%
%
%

\author{Jean-Rassaire Fouefack,
        Bhushan Borotikar,
        Tania S. Douglas, 
        Val\'erie Burdin,
        and~Tinashe~ E.M.~Mutsvangwa,~
\IEEEcompsocitemizethanks{\IEEEcompsocthanksitem Jean-Rassaire Fouefack, Tania Douglas and Tinashe Mutsvangwa are with the Division of Biomedical Engineering,
        University of Cape Town, 7935, South Africa.\protect\\
E-mail: ffcjea001myuct.ac.za, tania.douglas@uct.ac.za,\protect\\ tinashe.mutsvangwa@uct.ac.za
\IEEEcompsocthanksitem Jean-Rassaire Fouefack, Val\'erie Burdin  are  with the IMT-Atlantique and the Laboratory of Medical Information Processing (LaTIM INSERM U1101), Brest, France.\protect\\
E-mail: valerie.burdin@imt-atlantique.fr
\IEEEcompsocthanksitem Bhushan Borotikar is  with the Laboratory of Medical Information Processing (LaTIM INSERM U1101), CHRU de Brest, Brest, France and University of Brittany Occidental, Brest, France.
\protect\\
E-mail: bhushan.borotikar@gmail.com 
}
\thanks{This work is based on research supported by the National Research Foundation (NRF) of South Africa (grant no's 105950 and 114393); the South African Research Chairs Initiative of the NRF and the Department of Science and Technology (grant no 98788); the South African Medical Research Council and the ``Mission Enseignement Sup\'erieur, Recherche et Innovation'', Brest M\'etrople, France (grant no 17-178)}}

\IEEEtitleabstractindextext{%
\begin{abstract}
\justifying
Statistical shape models (SSMs) are state-of-the-art medical image analysis tools for extracting and explaining features across a set of biological structures. However, a principled and robust way to combine shape and pose features has been illusive due to three main issues: 1) Non-homogeneity of the data (data with linear and non-linear natural variation across features), 2) non-optimal representation of the $3D$ motion (rigid transformation representations that are not proportional to the kinetic energy that move an object from one position to the other), and 3) artificial discretization of the models. In this paper, we propose a new framework for dynamic multi-object statistical modelling framework for the analysis of human joints in a continuous domain. Specifically, we propose to normalise shape and dynamic spatial features in the same linearized statistical space permitting the use of linear statistics; we adopt an optimal 3D motion representation for more accurate rigid transformation  comparisons; and we provide a 3D shape and pose prediction protocol using a Markov chain Monte Carlo sampling-based fitting. The framework affords an efficient generative dynamic multi-object modelling platform for biological joints. We validate the framework using a controlled synthetic data. Finally, the framework is applied to an analysis of the human shoulder joint to compare its performance with standard SSM approaches in prediction of shape while adding the advantage of determining relative pose between bones in a complex. Excellent validity is observed and the shoulder joint shape-pose prediction results suggest that the novel framework may have utility for a range of medical image analysis applications. Furthermore, the framework is generic and can be extended to n$>$2 objects, making it suitable for clinical and diagnostic methods for the management of joint disorders.
\end{abstract}

}

\maketitle

\IEEEdisplaynontitleabstractindextext

%
\IEEEpeerreviewmaketitle

\IEEEraisesectionheading{\section{Introduction}\label{sec:introduction}}

%
%
%
%

\IEEEPARstart{F}{inding} a minimum subset $\mathbf{D}_0$ with $N_s$ patterns that can generally and specifically represent a given  set of structures $\mathbf{D}$  with $N$ feature classes is a frequently researched area \cite{c33}. A common strategy is to estimate a subspace of homogeneous patterns (for example, geometric shapes) through models. This inductive learning approach leads to models that allow for automated observation and interpretation of the learned patterns. In the medical image analysis domain, pattern analysis has to contend with two existing problems. First, patterns may present across non-homogeneous feature classes. An example is morphofunctional analysis where determining the changes in relative spatial dynamics of different anatomical structures adds an additional pattern class on top of the morphological (geometric) pattern class \cite{c31}. Second, patterns in some feature classes may belong to non-linear spaces. The current trend is to formalise the problem as an inverse problem and use deep learning based generative models \cite{deeppose1,deeppose2}. However deep generative models tend to have high specificity but low generality; in some cases explaining out-of-training data inputs with high confidence while being wrong \cite{overfitsolution1,overfitsolution2,deepmodelknowwhatdontknow}. Furthermore deep generative models require large numbers of training examples to be successful. 
 Finally, the interpretability of deep generative models remains complex. Thus, non-parametric analysis of non-linear variations of non-homogeneous features for different medical image analysis tasks (shape completion, morphometric analysis, segmentation) with the same model, remains desirable. 

With regard to analysis of shape, a well established and understood formalism for analysing 3D geometric variation in a linearized statistical space exists in the form of statistical shape modelling (SSM). SSMs typically model the data in Euclidean vector space using principal component analysis (PCA) that treats feature variation changes as a linear combination of local translations only \cite{c2,c3,SSMTinashe,c15}. While limited in capturing the non-linearity in shape space, the validity of this linearization for rigid shapes has been codified in the literature for single anatomical structures \cite{c2,c3,c15}. Efforts for faithfully representing the non-linearity of shape space have been reported but have not become mainstream due to computational inefficiency and a lack of robustness \cite{diffcoordinatesSSM}. Recently, \textit{Luthi et al (2018)}\cite{c15} introduced a generalisation of SSMs, referred to as Gaussian process morphable models (GPMMs). In this framework, the parametric low-dimensional model was represented as a Gaussian process over deformation fields obtained from training examples. In contrast to discrete models that are dependent on artificial discretization, GPMMs are inherently continuous, that is, permitting of the arbitrary discretization of the domain on  which the model is defined. However, GPMMs do not embed inter-object shape correlation, nor the anatomo-physiological relationship between articulating objects. This is because current SSMs and GPMMs are unable to represent, in a statistically robust and intuitive way, an articulating anatomical complex composed of several rigid substructures which can move relative to each other. 

 Similar to the shape case, pose variation using both linear and non-linear statistical descriptions have been reported \cite{c6,c18,c9,c22,c8,c4,c34}. Previous studies have reported the embedding of relative pose variation of articulating bones into SSM frameworks, in the form of statistical shape and pose models (SSPM) \cite{c6,c18}. Others have reported statistical pose models (SPM) only; that is, the model of variation across spatial orientations without inclusion of their shapes \cite{c5}. A straightforward approach to model shape and pose together is to model joint flexibility implicitly by incorporating joint motion in statistical space \cite{c22}. However, this assumes that pose variations are a statistical property of anatomy. Another approach is to concatenate pose and anatomical features into a joint vector and apply linear PCA to obtain joint principal components \cite{c9}. However, this  describes the pose transformations in Euclidean vector (linear) space; contrary to reality as they belong to a manifold and thus need to be described using non-linear statistical methods. 
 
 Various efforts have been reported in the literature to model shape and pose features together. \textit{Bossa et Olmos (2007)} proposed parametric low-dimensional models of several subjects in different poses with the pose variation representation obtained through principal geodesic analysis (PGA); a non-linear extension of PCA \cite{c34}. Shape and pose features of the brain cortex were concatenated in a long vector and standard multivariate statistics were extracted \cite{c18}. \textit{Fitzpatrick et al (2011)} \cite{c8} reported on a characterisation of the relationship between shape and contact mechanics of the patella and femur using an SSM of the articulating surfaces between both bones. The authors considered the patello-femoral contact space as a new object and modelled it as a standard SSM. \textit{Chen et al (2014)} proposed an articulated SSM for segmentation of the wrist from computed tomographic (CT) images without concatenating the shape and pose features; their model was a combination of an SSM and an SPM developed using PGA \cite{c7}. All the above shape and pose PGA based models used a standard representation (Rodrigues, quaternion and Euler representation) for modelling the rigid transformations describing the pose. \textit{Moreau et al (2017)} reported that the standard representation is limited in its ability to model rigid transformations of non-compact objects (with respect to a shape compactness measure). They instead proposed a new norm for statistical analysis of rigid transformations that is robust for analysing non-compact objects. However, they only demonstrated a new norm for SPMs without considering shape and pose variability analysis. A non-linear shape modelling approach based on differential coordinates (thus avoiding global rigid transformation through local rotation and stretching) has been reported \cite{surftheorySSM,diffcoordinatesSSM}. While this approach captures the inherent non-linearity in shape variation, it still suffers from artificial discretization which prevents marginalization of the resultant probabilistic model. Additionally, it does not allow for modelling multi-object structures.
 
 We hypothesise that for a principled approach to multi-object shape and pose analysis, the following limitations identified in the literature need to be addressed : 1) features of different classes (morphology and pose) need to be properly normalised to obtain an homogeneous analysis space; 2) $3D$ motion requires an optimal representation to induce an efficient associated norm for comparison of rigid transformations and 3) models need to be built in a continuous domain. We propose a novel framework that addresses all the above limitations by modelling shape and pose variation of complex articulated objects composed of multiple rigid objects. We refer to the proposed modelling framework as Dynamic Multi-object-Gaussian Process Modelling (DMO-GPM)\footnote{The code for this project is available online for research use: https://github.com/rassaire/DmoGpm}.

The structure of the rest of the paper is as follows. In section 2, we identify and describe previous and related work in the domain of SSPMs and GPMMs. In section 3, we describe the DMO-GPM framework and explain 1) how the framework is extended and generalised to a joint consisting of $N$ objects and the spatial positions are projected in a linear space using PGA, 2) how the new framework addresses the inconsistent representation of $3D$ motions typical in modelling joints with non-compact structures such as long bones using the standard representation, and 3) how we adopt an energy displacement representation that is more suitable for reconstruction of the objects of interest. In section 4, we validate the framework using synthetic data with predefined and accurately known shape and spatial dynamics. We also propose an automatic pipeline for dynamic multi-object prediction using a combination of Markov Chain Monte Carlo sampling and DMO-GPM approach which allows for registering a model to pathological or incomplete multi-objects while avoiding local minima. In section 5, we apply the framework to complex shoulder joint data to demonstrate the utility of the approach compared to the state-of-the-art sample kernel-based GPMM.

\begin{figure*}[thpb]
	\centering
	\includegraphics[width=1 \textwidth, angle =0 ]{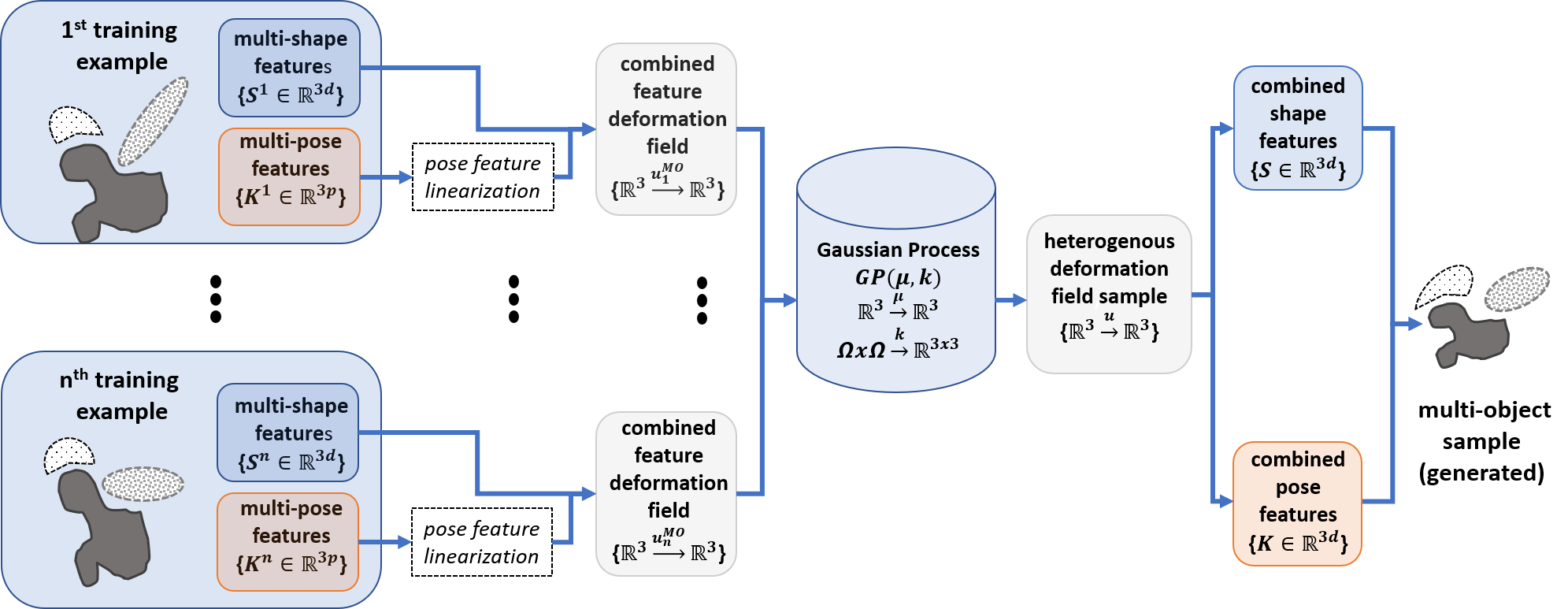}
	\caption{Proposed multi-object generative model architecture. Multi-shape and pose features are obtained after establishing anatomical correspondence. The pose feature linearization is performed through tangent space projection using $\log$ mapping.  A Gaussian process is defined over  deformation fields obtained from combined shape and pose features. Multi-object samples are obtained by 1) using the shape deformation field component to instantiate shape instances; and 2) a retrieval of their spatial orientation using rigid transformations associated to the pose deformation field component.} 
	\label{best_fit_scapula_humerus}
\end{figure*}

\section{Previous and related work}
Before describing our DMO-GPM framework, we summarise the main concepts behind traditional statistical shape models, multi-object statistical shape and pose models, and GPMMs on which we will capitalise to synthesize the dynamic multi-object framework.
\subsection{\textbf{Statistical shape and pose models}}
 Let's consider a multi-object data-set $\{\{S_i^j\}_{i=1\dots n}\}_{j=1\dots N}$ with $n$ examples, where each example is composed of $N$ objects. For example, the gleno-humeral joint with training data samples and having $2$ rigid bony objects - scapula and humerus. The underlying assumption in the development of multi-object statistical shape and pose models (SSPMs) is that there exists a representational space where the relationship between shape and pose can be learnt from the data-set. This concept of SSPMs is well-adapted for modelling articulated objects in motion.
The $j^{th}$ object of the $i^{th}$ data sample can be represented by $S_i^j$, a vector of discrete landmarks with known correspondence between the examples $\{S_i^j\}_{i=1\dots n}$ of the $j^{th}$ object, the reference object $\Gamma_S^j$ and $\vec{\Gamma}_S^j$ a coordinate vector  of points of $\Gamma_S^j$, typically, one selected object from the database that should be closest to the mean.
The rigid transformation that aligns the $j^{th}$ object of the $i^{th}$ example to its reference object $\Gamma_S^{j}$ is defined as
\[ h^j_i=\underset{ h\in SE(3)}{\arg min} \|h(S^{j}_i)-\vec{\Gamma}_S^{j}\|^2\]
where $SE(3)$ is the space of rigid transformations defining a Lie group with $\|.\|$ being the sum of distance metric. The Lie group (smooth manifold) defined by the system of objects is the direct product of the $N$ rigid transformation groups denoted as $SE(3)^N$. Shape variations are relatively small and are traditionally assumed to belong to a vector space, while pose (rigid transformations) variations belong to a Riemannian manifold which is not a vector space. In contrast to shape models that are built using a classical PCA, the pose models are built using PGA which is in essence a PCA in a space tangential to the manifold\cite{c34,c21}. Several studies use PGA to perform linear statistics on a Riemannian manifold \cite{c34,c7,c6,c4}. To perform the PGA, the $i^{th}$ transformation for the $j^{th}$ object is represented in this tangent space to $SE(3)$ with respect to the mean rigid transformation $\bar{h}_j$ using the log mapping:
\begin{equation}\label{eurler parm}
\begin{aligned}
log_{\bar{h}_j}h_i^j=[p_1,\dots,p_m]^T
\end{aligned}
\end{equation}

where $(p_i)_{i=1\dots m}$ are the parameters of the rigid transformation $\bar{h}_j^{-1}\circ h_i^j$ representing the Euler's, Rodrigues's or quaternion parameters. These representations describe the $3D$ rotations in 3D. Rodrigues and quaternion  represent a $3$D rotation as an axis vector and a rotation angle while Euler represents it with three angles. For the rest of the paper these representations will be called standard representations (SR). To map the $i^{th}$ transformation back to the manifold, an exponential mapping is used:
\[ h_i^j=\bar{h}_jexp(log_{\bar{h}_j}h_i^j)\]
To build the SSPMs, shape and pose features of the $i^{th}$ example are concatenated in joint feature vectors \cite{c18} by:
\[X^i= [w^sX^i_S,w^pX^i_p]^T\]
where $X^i_S=[S_i^{1},\dots, S_i^{N}]^T$ and $X^i_p=[log_{\bar{h}_j}h_i^1,\dots,log_{\bar{h}_j}h_i^N]^T$, $w^s$ and $w^p$ denote shape and pose weightings, respectively, that balance the relative importance between shape and pose within the model. The resulted norm of the SR is defined as:
\begin{equation}\label{pdms}
\begin{aligned}
\Vert h_i^j \Vert_{SR}=t^T\; t+s\, r^T\,r
\end{aligned}
\end{equation}
where $r$ and $t$ are rotation and translation vectors respectively; and $s$ the scaling factor.
To build the model, an additional assumption is that the joint feature residual data-set $\{X^i\}_{1\dots n}$ can be modelled by a normal distribution similar to point distribution models PDMs \cite{c2} as follows:
\[(s,p)\sim \mathcal{N}(\mu,\Sigma)\]

where $\mu$ is the joint mean and $\Sigma$, the covariance matrix, are estimated from the data-set $\{X^i\}_{1\dots n}$ as:

\[\mu=\bar{X}=n^{-1}\sum_{i=1}^{n}X^i\]
\[\Sigma=\bar{X}=n^{-1}\sum_{i=1}^{n}(X^i-\bar{X})(X^i-\bar{X})^T\]
The model is obtained by performing PCA which in a probabilistic interpretation leads to:

\[X=\bar{X}+\sum_{k=1}^{n}\alpha_k\sqrt{\lambda_k}\, e^k\]

where $e^k=[e^k_S,e^k_p]^T$ is the combined $k^{th}$ eigenvector of the covariance matrix (associated to the  $k^{th}$ eigenvalue $\lambda_k$), obtained from single value decomposition (SVD) and $\alpha_k\sim \mathcal{N}(0,1)$.
Then, the SSPM is split into two parts, the shape component, typically a PDM, defined as follows:
\begin{equation}\label{pdms}
\begin{aligned}
X_{\alpha_k}=\bar{X_s}+\sum_{k=1}^{n}\alpha_k\sqrt{\lambda_k}\, e^k_s 
\end{aligned}
\end{equation}

and the pose component, typically SPMs, defined as:

\begin{equation}\label{modefromation model}
\begin{aligned}
 h_{\alpha_k} = \bar{h}\, exp\left ( \sum_{i=1}^{n}\alpha_k\sqrt{\lambda_k}e^k_p \right )
\end{aligned}
\end{equation}
A new instance of a discrete  SSPM is obtained as $s_p=  h_{\alpha_k}(X_{\alpha_k})$ which derives the shape at a relative spatial position. 

\subsection{\textbf{Gaussian process morphable models}}

Gaussian process morphable models\cite{c15} are an extension of PDMs in equation \ref{pdms}. In contrast to PDMs, GPMMs model deformations defined on a continuous domain, although they can always be discretized to obtain a model that is mathematically equivalent to a PDM. The advantage of such model is that after training, the statistical descriptions may be marginalized (or customized kernels could be used)  to a region of interest which will provide better interpretability of the underlying information. Additionally, having freedom in defining the covariance function may be useful because explicit bias kernels can be added based on expert knowledge. This GPMM flexibility overcomes a typical limitation of traditional SSM whereby insufficient training examples result in loss of model generality \cite{c32}. 
In GPMMs, shape correspondence, established across $n$ examples, leads to deformation fields $u_i, i=1\dots n$ defined from the reference shape $\Gamma_S$ to $i^{th}$ examples in the training data. These deformations are then modelled as Gaussian Processes ($GP$) on the reference shape domain $\Omega_S$ ($\mathbb{R}^3 \supset \Omega_S \supset\Gamma_S$) thus making them independent of discretization. The GPMMs are formulated as follows:

\begin{equation}\label{gp model}
\begin{aligned}
\{\Gamma_R+u(\Gamma_R)\} \text{ with } u\sim GP(\mu, k)
\end{aligned}
\end{equation}
where for all points $x,y$ of $\Omega$, the mean and covariance (kernel function) are defined as:
\begin{equation}\label{mean and kernel of GPMM}
\begin{aligned}
&\mu(x)= n^{-1}\sum_{i=1}^{n}u_i(x)\\
&k(x,y)=n^{-1}\sum_{i=1}^{n}(u_i(x)-\mu(x))(u_i(y)-\mu(y))^T
\end{aligned}
\end{equation}
  The function $u_i$ is the deformation field associated with the dense correspondence established across the shape parameters between the reference and the $i^{th}$ example.  The deformations can be represented in terms of an orthogonal set of basis functions $\{\Phi\}_{m=1}^{M}, M\le n$:
\begin{equation}\label{spms}
\begin{aligned}
u(x)=\sum_{m=1}^{M}\alpha_m\sqrt{\lambda_m}\Phi_m(x), \text{ with } x\in \Omega_S, \, \alpha\sim \mathcal{N}(0,1)
\end{aligned}
\end{equation}
where $(\lambda_m, \Phi_m)$ are the eigenvalue/eigenfunction couples. These couples are obtained using Karhunen-Lo\`eve expansion and Nystr\"om approximation \cite{c28,c29}, as explained in \cite{c15}. That leads to a probabilistic PCA-based  model of shape directly on the deformations. 

\section{Dynamic multi-object Gaussian process models}
In this section we develop the dynamic multi-object Gaussian process model (DMO-GPM) framework that incorporates:
\begin{itemize}
    \item an homogeneous analysis space (linear) for shape and spatial orientation features,
    \item a kinetic energy based $3D$ motion representation for more accurate comparison of rigid transformations,
    \item a continuous representation to describe feature variation across the training dataset.
\end{itemize}

The DMO-GPM framework is aimed at modeling multi-objects (with $N$ objects) and their relative spatial positions as a deformation field  $u^{MO}=\{u^j_S,u^j_K\}_{j=1}^N$ from a reference joint $\Gamma=\cup_{j=1}^{N}(\Gamma^j_S\cup\Gamma^j_K)\subset\mathbb{R}^{3}$, where a $j^{th}$ object shape of the joint, at its relative spatial position, is represented as
\begin{equation}\label{bref decription of DMO-GPM}
h^j(\{x+u^j_S(x)|x\in\Gamma^j_S\})
\end{equation}
with the rigid transformation $h^j$ defined as
$$h^j=\underset{ h\in SE(3)}{\arg min} \|\vec{u}_h-\vec{\Gamma}_K^{j}\|^2$$
for some deformation $u^{MO}:\Omega=\cup_{j=1}^{N}(\Omega^j_S\cup\Omega^j_K)\subset\mathbb{R}^3\rightarrow \mathbb{R}^3$, with $\Gamma\subset\Omega$ and $\vec{u}_h$ being the coordinate vector of the set of points $h(\{x+u^j_K(x)|x\in\Gamma^j_K\})$. We model deformations as a Gaussian process $u^{MO}\sim GP(\mu,k)$ with mean function $\mu:\Omega\longrightarrow\mathbb{R}^d$ and covariance (kernel) function $K:\Omega\times\Omega\longrightarrow\mathbb{R}^{3\times 3}$.

The DMO-GPMs are built through 4 steps. First, the anatomical correspondence needs to be established across the multi-objects to guarantee the homology of shape-pose features. Second, the optimal representation of spatial dynamics associated with shapes is obtained. Third, linear and non-linear transformations are normalized to obtain an homogeneous space for linear statistics. Finally, the dynamic multi-object is modelled in a continuous domain.

\subsection{Multi-object correspondence establishment}\label{correspondence}
We define a dynamic multi-object (joint) as a set of several biological structures (or objects) from different family of shapes (e.g humerus, scapula etc.) and their relative spatial position. Establishing correspondence across several structures while keeping their anatomo-physiological relationship is not a straight-forward task. Here we extend the framework presented in our previous work \cite{c14} for two biological structures to $N$ ($N\in \mathbb{N}$) biological structures. For a single biological structure, amongst several different approaches \cite{c25,c26,c27}, correspondence can typically be achieved through rigid-registration and subsequently non-rigid registration performed using free-form deformation (FFD) models such as GPMMs \cite{c15}. For multi-objects, not only shape correspondence should be established across objects, but also the relative spatial position between them should be preserved. To register $N$ objects while keeping their anatomo-physiological relationship, we assume that one of the $N$ objects remains fixed while the others move relative to the fixed one. Individual FFD models ($N$ models) are built for each object family and used to register objects of their respective family. After the registration, each inverse of the rigid registration transformation is used to retrieve the original spatial position of moving objects. This operation retrieves the anatomo-physiological relationship between moving objects. Rigid transformation of the fixed object is finally applied to all moving objects to obtain their anatomo-physiological relationship with the fixed object. Dense correspondence (shape and spatial dynamics) is established across the data-set by using the same FFD as reference for the registration.
 
 Let $T_j$ be the rigid transformation that transforms the $j^{th}$ object ($S^j, j=1\dots N-1$) to the reference $\Gamma_S^j$, $T$ the rigid transformation that transforms the fixed object ($S^{j_0}$) to its reference $\Gamma^{j_0}$. The $j^{th}$ registered object, $S_r^j$, with its spatial position retrieved is obtained as:
 
$$S_r^j =T\circ T_j^{-1}( \underset{u \in DF}{\arg min} SM(u(\Gamma_S^j),S^j)+\beta\|u\|^2)$$
where $DF$ is the space of deformation fields, $\|.\|$ is a norm in $DF$, $SM$ is a similarity measure and $\beta$ a regularisation coefficient.

\subsection{Optimal representation of $3D$ motion}
The literature on SSPMs usually emphasizes on the common shape-pose space required in medical image processing. The standard representation of rigid transformations may not always encapsulate the kinetic energy necessary to move points on an object from one position to another, particularly for non-compact shapes. This reduces the efficiency of this parameterization for statistical analysis of rigid transformations \cite{c4}. Thus, we propose to extend the kinetic energy based representation by \textit{Moreau et al} (2017) in a continuous space and use it to encode a rigid transformation in DMO-GPM. We refer to this new transformation representation as the energy displacement representation (EDR).

We represent a multi-object complex as a set of $N$ objects; each object defined by its shape and its spatial position parameters. A multi-object is represented as a concatenated vector space and denoted as $X= [X_S,X_K]^T$ where $X_S=[S^{1},\dots, S^{N}]^T$ and $X_K=[K^1,\dots,K^N]^T$ with $S^{i}$ and $K^{i}$ representing the point domain of the shape and the spatial orientation features for the $i^{th}$ object of the joint, respectively.


Now let us assume there are $n$ multi-objects examples in the training data-set. Let us explicitly define the rigid transformation representation which is the displacement field representation that extends \cite{c4} in a continuous domain and will be used to define joint features representations. We define the EDR of the $i^{th}$  example of the object $j$ on its reference spatial position domain $\Omega_{K}^j\subset\mathbb{R}^3$ with their respective reference shape $\Gamma^j_S$ and reference pose points $\Gamma_K^j$. The EDR $\delta^{j}_i$ and its associated metric $d$ are defined as:

\begin{equation}\label{EDR}
\begin{aligned}
&\delta^{j}_i(x)=h^j_i(x)-Id(x), x \in \Omega_{K}^j, i=1\dots n \\
& h^j_i=\underset{ h\in SE(3)}{\arg min} \|h(S^{j}_i)-\vec{\Gamma}_S^{j}\|^2\\
&d^2(h^j_i,Id)=\delta^{j}_i(\Omega_{K}^j)^T\delta^{j}_i(\Omega_{K}^j)
\end{aligned}
\end{equation}
where $SE(3)$ is a lie group of rigid transformations.

\subsection{Normalising dynamic multi-object features for homogeneous representation}
To normalise shape and pose features, we define unified representations of shapes and their relative spatial dynamics. These representations are joint deformation fields over which a Gaussian process (GP) will be defined (section 3.4). A reference multi-object $\Gamma=\cup_{j=1}^{N}(\Gamma^j_S\cup\Gamma^j_K)\subset\mathbb{R}^{3}$ is chosen from the training examples. This joint should be representative of the data-set, that is, approximately similar to the mean of the data-set. A joint deformation field of the $i^{th}$ example in the training data-set is a pair of deformation fields ($u^i_S$, $u^{i}_K$), where $u^i_S$ maps a point $x$ of the reference shape objects $\cup_{j=1}^{N}\Gamma^j_S$ to the point $u^i_S(x)$ of its corresponding shape example; and $u^i_K$ maps a point $x$ of the reference pose objects $\cup_{j=1}^{N}\Gamma^j_K$ to the point $u^i_K(x)$ of its corresponding pose example. The $i^{th}$ joint deformation field is defined 
\begin{equation}
 \begin{dcases}
 u^i_S(x) \text{ if } x\in \Omega_S=\cup_{j=1}^{N}\Omega^j_S,\\
  u^i_K(x) \text{ if } x\in \Omega_K=\cup_{j=1}^{N}\Omega^j_K.\\
\end{dcases}
\end{equation}
where 
\begin{equation}
\begin{dcases}
u^i_S(x)= u^j_i(x), \text{ if } x\in\Omega_S^j\\
        u^i_K(x) =h^j_i(x),\text{ if } x\in\Omega_K^j.\\
\end{dcases}
\end{equation}

Before obtaining residual features (around the mean) that will represent normalised features, the joint function of the mean shape and spatial transformations has to be defined. We estimate this mean function, $\mu_{MO}$, using the mean pose deformation fields as:
\begin{equation}
\begin{dcases}
\frac{1}{n}\sum_{i=1}^{n}u^j_i(x), \text{ if } x\in\Omega^j_S,\\
        \mu^j_K(x)\approx \bar{h}^j(x),\text{ if } x\in\Omega_K^j.\\
\end{dcases}
\end{equation}
with
$\bar{h}^j=\underset{ h\in SE(3)}{\arg min} \Big\|h\Big[\frac{1}{n}\sum_{i=1}^{n}\delta^{j}_i(\vec{\Gamma}_K^{j})\Big]-\vec{\Gamma}_K^{j}\Big\|^2$.\\
It should be noted that the mean $\bar{h}$ is obtained using the mean of the vector field (thanks to EDR), in contrast to SR where the  Fr\'echet mean is estimated through numerical optimization.

Thus far, pose representations across the training data-set belong to a manifold which is a non-linear space. Rigid transformations are projected (that is linearization of relative spatial transformations) onto a linearized space to the manifold at the mean rigid transformation (tangent space) using the EDR (eq \ref{EDR}), in order to obtain the unified space of shape and pose features. Relative spatial transformations  are linearized  through $\exp/\log$ bijective mapping.

Let us define the $\exp/\log$ functions associated to the EDR between two rigid transformations $h^j_{i_1}$, $h^j_{i_2}$:
\begin{equation}
\begin{aligned}
&\log[h^j_i]=\delta_i^{j}, j=1\dots N \text{ and } i=1\dots n\\
&\log_{h^j_{i_1}}[h^j_{i_2}]=\log[(h^j_{i_2})^{-1}\circ h^j_{i_1}]\\
&\exp[\delta^{j}_i]=\underset{ h\in SE(3)}{\arg min} \|h[\delta^{j}_i(\vec{\Gamma}_K^{j})]-\vec{\Gamma}_K^{j}\|^2\\
&\exp_{h^{j}_{i_1}}[\delta^{j}_{i_2}]=h^{j}_{i_1}\circ \exp[\delta^{j}_{i_2}].
\end{aligned}
\end{equation}

The projected joint deformation fields onto the vector space over which the GP will be defined are $\{u_i^{MO}\}_{i=1}^n$
\begin{equation}
u_i^{MO}=\begin{dcases}
u^i_S(x)-\mu_{MO}(x), x\in \Omega_S\\
        \log_{\mu_{MO}}[u^i_K](x), x\in\Omega_K.
\end{dcases}
\end{equation}

\subsection{Dynamic multi-object modeling in a continuous domain}\label{GP multi-object}
From a training dataset of $n$ examples with $N$ objects each, the deformation fields $\{u_i^{MO}\}_{i=1}^n$ defined on the continuous domain $\left ( \Omega=\cup_{j=1}^{N}(\Omega^j_S\cup\Omega^j_K) \right )$ are modelled by a Gaussian process defined as:
\begin{equation}
u\sim GP(\mu_{MO}, k_{MO}) 
\end{equation}
where  $k_{MO}$ is the kernel function defined as:

\begin{equation}
k_{MO}(x,y)=\frac{1}{n}\sum_{i=1}^{n}u^{MO}_i(x)u^{MO}_i(y)^T, x, y\in \Omega\\
\end{equation}
 The multi-object deformation fields $u^{MO}$ are represented by an orthogonal set basis functions denoted by $\{\phi_m^{MO}\}_{m=1}^{M}$, $M\le n$ and defined as:
 \begin{equation}
 \phi_{m}^{MO}(x)=\begin{dcases}
  \phi_m^S(x) \text{ if } x\in \Omega_S,\\
        \phi_m^K(x) \text{ if } x\in \Omega_K.\\
\end{dcases}
\end{equation}
where $\phi_m^S$ and $\phi_m^K$ represent the shape and the energy displacement component of the $m^{th}$ basis function.

The $j^{th}$ object is modelled as a combination of shape deformation field and a rigid transformation (obtained using exponential map) as defined below: 
\begin{equation}
\begin{dcases}
 \sum_{m=1}^{M}\alpha_m\sqrt{\lambda^{MO}_{m}} \phi_{m}^{S_j}(x), \text{ if } x\in\Omega^j_S,\\
  h^j:\exp_{\mu_{MO}}\left(\sum_{m=1}^{n}\alpha_m\sqrt{\lambda^{MO}_{i}} \phi_{m}^{K_j}(x)\right),\text{ if } x\in\Omega_K^j.\\
\end{dcases}
\end{equation}
where $\alpha\sim \mathcal{N}(0,1)$ and the pair $(\lambda_{m}^{MO}, \phi_{m}^{MO})_{m=1\dots M}$ representing  the $m^{th}$  eigenvalue and eigenfunction couple. These couples are obtained using Karhunen-Lo\`eve expansion and Nystr\"om approximation as in equation \ref{spms}. The $m^{th}$ eigenfunction $\phi_{m}^{MO}$ is a continuous multi-object deformation field defined on $\Omega$, in the tangent space to $SE(3)$.

\begin{figure}[thpb]
	\centering
	\includegraphics[width=0.45 \textwidth, angle =0 ]{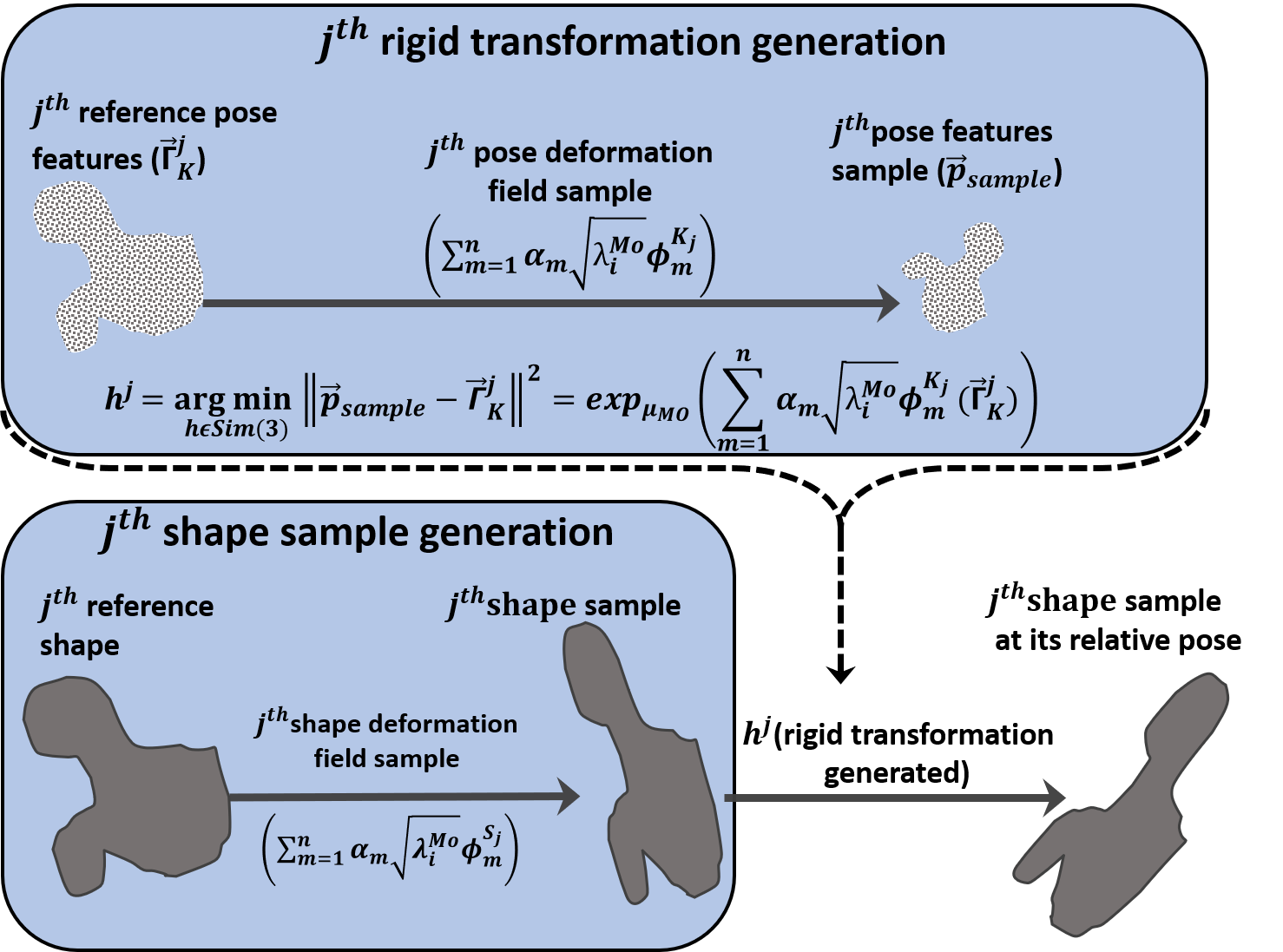}
	\caption{Obtaining shape sample at its associated spatial orientation. Top: the rigid transformation is obtained from pose deformation filed sample. Bottom: the sample sample is obtain from shape deformation filed and its spatial orientation is retrieved using the rigid transformation generated } 
	\label{samplegeneration}
\end{figure}

  For example, figure \ref{samplegeneration} shows how a $j^{th}$ object is sampled from the model.  A shape instance of the $j^{th}$ object and its associated spatial position can be generated from our probabilistic model for $\alpha=(\alpha_m)_m\sim \mathcal{N}(0,1)$ as:

\begin{equation}
\begin{aligned}
& \textrm{shape\&pose}^j=h^j(\textrm{shape}^j)\\
&  \text{with}\\ 
& \textrm{shape}^j=\sum_{m=1}^{M}\alpha_m\sqrt{\lambda^{MO}_{m}}\phi_{m}^{S_j}(\Gamma_S^j)
\end{aligned}
\end{equation}

where $\Gamma_S^{j}\subset\Omega_S^j$ is the reference shape of the $j^{th}$ object.
\subsubsection{Marginalising the model}
As in GPMMs, our DMO-GPM framework allows to operate on a specific part of the domain for a given model using marginal probability properties, DMO-GPMs still have these properties. 

Let us assume that the DMO-GPM is a joint probabilistic model denoted by $p_{\textrm{MO}}(\Omega)$.
The marginal probabilistic model is defined as 
\begin{equation*}
\begin{aligned}
&p_{\textrm{MO}}(\Omega_i):= (\mu_{MO}, k_{MO})|_{\Omega_i \subset \Omega},\\
\end{aligned}
\end{equation*} 
obtained from the joint probability in a specific domain. 

Another important aspect of the DMO-GPM framework is its potential of making the model ``class specific'', that is, making the variations across a specific class constant. For example, this model includes static multi-object models, that is, setting the pose parameters constant to obtain a classic SSM of the joint. Conversely, it includes a motion model, that is, setting the shape parameters constant which leads to a pose-only statistical model.

\subsubsection{Modelling with kernels}
Another advantage of the  DMO-GPMs is the possibility of increasing the variability around the mean without retraining the model as in GPMMs. The number of training examples may be insufficient to provide a subspace of shapes to represent all possible shape variations within the family, which introduces bias. The extension of the variability using a bias kernel can overcome this limitation. 
Our framework allows for the understanding of such variability and defining an appropriate kernel that encodes the variability. A new kernel can be added to the trained model kernel without a need of reusing the training examples. An example of kernel that can increase smoothness in a model is the Gaussian kernel defined in a specific domain $\Omega^j_S \text{ or } \Omega^j_K,j=1,\dots,N$ by

\begin{equation*}
k_{\Omega}^{(s,\sigma)}(x,y)=s\exp(\|x-y\|^2/\sigma^2),x,y\in \Omega
\end{equation*} 
where the parameter $s$ defines the scale of the average error and where $\sigma$ the range over which deformations are
correlated \cite{c15}. The new DMO-GPM kernel can be a combination of the pre-trained kernel with different Gaussian kernels in each domain defined by
\begin{equation*}
k_{new}= k_{\textrm{MO}}+\sum_{j=1}^{N}I_{3\times 3}k_{\Omega_j}^{(s_j,\sigma_j)}, \Omega_j\in \{\Omega^j_S, \Omega^j_K\}
\end{equation*} 
where the $I_{3\times 3}$ is a $3\times 3$ unit matrix.

\subsection{Using the framework in analysis tasks}
In order to demonstrate the utility of the framework in prediction from the observation of target examples, we adopted the Monte Carlo Markov chain (MCMC) and Metropolis Hastings algorithms as in \cite{c7,c32}. Markov Chain Monte Carlo methods are computational tools to perform approximate inference with intractable probabilistic models by making the use of another distribution from which one can sample, in order to ultimately draw samples from the true posterior distribution model. 

Let us consider target objects with arbitrary relative positions to each other. The goal of the MCMC fitting process is to find the best model parameters $\theta$ (a multi-object deformation field in our case) that optimally represents the target example. The uncertainties related to pose estimation are embedded in the DMO-GPM and therefore need not be modelled separately. The target multi-object example  may consist of partial objects; for instance, images of bones with slight cracks caused by osteomalacia or any incomplete anatomical structure on images due to pathology or due to image artifacts. The advantage of the DMO-GPM is that when confronted with a missing part on one object, the other objects in the target example can be used as additional observations since the correlation between the objects is embedded in the model. The DMO-GPM with MCMC fitting could be used to obtain a posterior model of one object while another is missing. 

Let $T_o$ be the observation (target multi-object complex), $T_o=\{T_o^j\}$ is a set of $N$ objects at various relative positions. Note that $T_o$ could be partial i.e. we may have less than $N$ objects. The posterior model is estimated using the Bayes rule:
\begin{equation*}
p(\theta|T_o)=\dfrac{p(T_o|\theta)p(\theta)}{p(T_o)}
\end{equation*} 
where $p(T_o)$ is intractable, $p(\theta)$ is the model and $p(T_o|\theta)$ the likelihood. To estimate the posterior DMO-GPM, a multi-object likelihood is proposed.

\subsubsection{Multi-object likelihood}
Full shape and pose estimations require a likelihood that captures within-object and between-object features. We propose the multi-feature class comparison likelihood that has a global and local component. The global component ensures the prediction of all feature classes while each local component ensures the predominance of its feature class in the global likelihood. 
The $j^{th}$ local likelihood $l_j$ compares the observation of the target $T_o^{j}$ with its instance $\theta(\Omega^{j})$ generated from the model using independent feature-wise comparison. It can be seen as a probability distribution of possible shape-pose samples evaluated for the target and it is defined as: 
\begin{equation}
l_j(\theta,T_o^j)=p(T_o^j|\theta(\Omega^{j}=\{X_i\}_1^P))=\prod_{x_i,x_j\in T_o^j}^Pp(x_j|\theta(x_i))
\end{equation}
The global likelihood $l$ compares the observation of target $T_o$ with the model instance $\theta(\Omega)$ and it is defined as: 
\begin{equation}
l(\theta,T_o)=p(T_o|\theta(\Omega=\{X_i\}_1^{PN}))=\prod_{x_i,x_j\in T_o}^{PN}p(x_j|\theta(x_i))
\end{equation}

\subsubsection{Obtaining the optimal parameter set $\theta$}
A sample $\theta^{'}$ is accepted as a new state $\theta$ for the $j^{th}$ object with the metropolis decision probability $\beta_j$ and the target $T$ with the probability $\beta$ defined by:

 \begin{equation}
 \beta_j=\min\left\{\dfrac{l_j(\theta, T_o^j)p(\theta)}{l_j(\theta^{'}, T_o^j)p(\theta^{'})}\dfrac{Q(\theta^{'}|\theta)}{Q(\theta^{'}|\theta)},1\right\}
\end{equation} 

\begin{equation}
 \beta=\min\left\{\dfrac{l(\theta, T_o)p(\theta)}{l(\theta^{'}, T_o)p(\theta^{'})}\dfrac{Q(\theta^{'}|\theta)}{Q(\theta^{'}|\theta)},1\right\}
\end{equation} 
where $Q(\theta^{'}|\theta)=\mathcal{N}(\theta^{'}|\theta, \Sigma_{\theta})$ is the proposal generator. Finally, the  proposal $Q$ is  fed  through  a sequence of  Metropolis acceptance decisions:
\begin{equation}
\{\beta_j, \beta\},j=1\dots,N
\end{equation}
The predicted target is the instance of the posterior model with the highest probability $\beta$.

\section{Validating the framework on synthetic data}
 To validate the DMO-GPM framework, we created synthetic data set with precisely defined modes of shape and pose variation. We avoided real biological data because these typically exhibit complex shape and pose variations making them difficult to use for bench-marking and validation. Our synthetic data consisted of surface mesh data of a "lollipop" as defined in \cite{c20}. The lollipop has a shaft, a neck and head. We used this template to generate  $30$ surface data with meshes composed of $6000$ vertices; all with identical surface topology. The only difference in shape between all generated meshes was the length of the major axis defining the ellipsoid of the head of the lollipop. This was varied uniformly in the range from $1~ mm$ to $−30 ~mm$, so the shape variation was entirely encoded using one degree of freedom. The shaft and neck areas were identical for all synthetic data.
 \subsection{Synthetic experiments with static objects}
\begin{figure*}[thpb]
	\centering
	\includegraphics[width=0.9 \textwidth, angle =0 ]{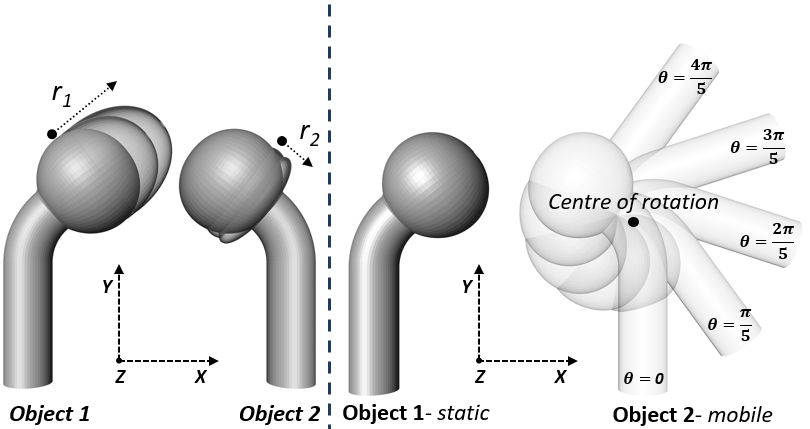}
	\caption{Illustration of the lollipop data.  Left) Shape variation of the first and the second objects, the first object parameter $r_1$ increasing in the direction of the arrow and the second object parameter $r_2$ decreasing in the direction of the arrow. Right) An example motion of a lollipop joint, the fixed object (grey) and the moved object (white) at different angle $\theta$.} 
	\label{lollipop_joint}
\end{figure*}

 To evaluate the potential of our framework in modelling shape correlation between different objects, we created artificial joints using two lollipops per scenario. Each joint was composed of two lollipops with major axes for corresponding pairs of lollipops of $r_1$ and $r_2$, for object 1 and object 2, respectively. The span of the data-set of joints was created by varying $r_1$ and $r_2$ as $(r_1,r_2)=(i/10,j/10),i=16,\dots,30, j=1,\dots,15$ creating a negative correlation between objects 1 and 2 (figure \ref{lollipop_joint} (left)).
\subsubsection{Empirical validation of correlation between objects}
The correlation between samples from the DMO-GPM and the one between the synthetically created examples were compared to know whether our model could establish the correlation between objects in the training data if it exists.
Our model  captured the correlation prescribed in the training data-set. The DMO-GPM showed a non-zero slope at $0.0001$ significance level with a negative correlation coefficient of $-0.99$ as in the training data-set. The regression curve of the training examples and our model examples were the same, illustrating the precision to which our model captured the correlation between objects of the synthetic joints.
 

 \subsubsection{Prediction using static model and MCMC}
 Prediction using statistical models remains challenging for medical image processing tasks because it requires observations that may not be available.  The models developed  have the potential to make predictions of one object without or with limited observations (which are the features available) of that object through leveraging information from the other object in a complex. The assumption is that there is a good correlation between the two objects. For the rest of the study we define four types of observations of target objects (see table \ref{table for type of obervations}) that will be used for predictions.
  \begin {table}
\caption {Taxonomy of observations of target object during predictions}
\label{table for type of obervations} 
 \begin{center}
    \begin{tabular}{ |p{1cm}| p{7cm}| }
    \hline
Type I & The full object to be predicted is available\\
\hline
Type II  & Partial object to be predicted is available \\
\hline
Type III  & Anatomical features (e.g. landmarks) of the object to be predicted are available\\
\hline
Type IV & No knowledge of the object to be predicted is available\\
 \hline
\end{tabular}
\end{center}
\end{table}
 
 To test this hypothesis, we generated $5$ joints of lollipops with the parameters $r_1$ and $r_2$ between the minimum and the maximum parameters that were in the training data-set. For each joint, object 1 was predicted from the observation of object 2 only and conversely. We compared the results obtained for each object with Type I observation to the ones obtained with Type IV observation. Table \ref{prediction lillipop using correlation} shows the root mean square (RMS) distance and the Hausdorff distance (HD) errors of the predictions. The average RMS error was $0.4\pm 0.06$ and $0.22\pm 0.02$~ mm for the prediction of object 1  and object 2 with Type I observation, respectively. The average HD error was $0.9\pm 0.22$ and $0.82\pm 0.2$ mm for the prediction of object 1 and object 2 with Type IV observation, respectively. These errors are relatively large compared to those predicted with Type I observation. However, this result obtained with type IV observation shows that the correlation embedded in the model provides additional information that can be useful in getting accurate prediction.
 \begin {table}
\caption {Prediction errors of the lollipop joint objects with Type I and IV observations using MCMC and DMO-GPM }
\label{prediction lillipop using correlation} 
 \begin{center}
    \begin{tabular}{ |p{4.5cm}|p{1.5cm}|p{1.5cm}|  }
    \hline
 & RMS (mm) &HD (mm)\\
 \hline
 Object 1 with Type I observation & \text{0.002}$\pm$\text{2.2e-6} &\text{0.014}$\pm$\text{1.7e-3}\\
         \textbf{Object 1 with Type IV observation} & \textbf{0.4}$\pm$\textbf{0.06}& \textbf{0.9}$\pm$\textbf{0.22} \\
  &  & \\
 Object 2 with Type I observation & \text{0.001}$\pm$\text{2.9e-6}&\text{0.01}$\pm$\text{0.001}\\
 \textbf{Object 2 with Type IV observation} &\textbf{0.22}$\pm$\textbf{0.02}&\textbf{0.82}$\pm$\textbf{0.2}\\
 \hline
\end{tabular}
\end{center}
\end{table}

 \subsection{Synthetic experiments with dynamic joints}
 To evaluate the potential of our framework in modelling shape and spatial dynamics together, we simulated joint motion. First, we created a data-set where shape does not correlate with motion. For each joint generated above, we rotated the second lollipop (object 2) relative to the first one (object 1) using the Euler's angle convention for describing  rigid transformations $(\varphi,\theta, \psi)$. The second lollipop was moved in the $yz-plane$ corresponding to Euler's angles $(0,\theta, 0)$; where only $\theta$ was varied by four angles ($\theta=\frac{1}{5}\pi,\frac{2}{5}\pi,\frac{3}{5}\pi,\frac{4}{5}\pi$) defining a motion (figure \ref{lollipop_joint} (right)). Defining the same motion for each joint created a data-set where shape was uncorrelated with motion because each pair of shapes performed the same movement. Secondly, we created a data-set where the motion is constrained by the shape. This was done by simulating the same movement as above but the maximum angle was inversely proportional to $r_2$ ( $\theta_{max}=-\frac{3}{4r_2}\pi$), thus creating correlation between shape and motion. 
 
 \subsubsection{Non-correlation and correlation of shape-pose variation}
\begin{figure}[thpb]
	\centering
	\includegraphics[width=0.48 \textwidth, angle =0 ]{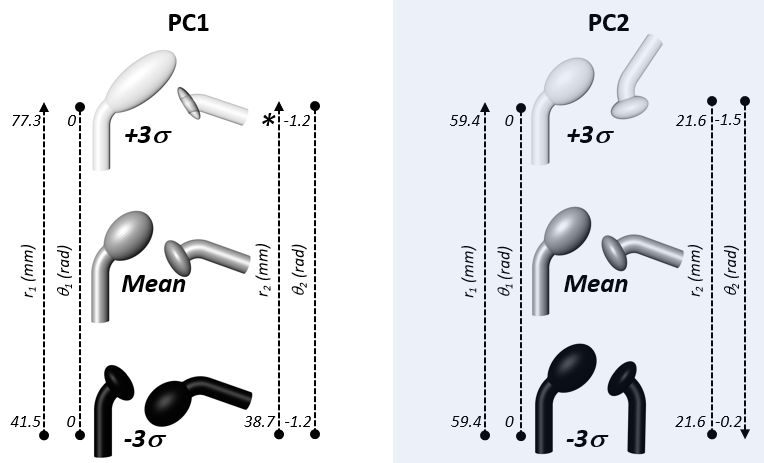}
	\caption{First and second PCs of the DMO-GPM with  no correlation between shape and  pose variation. The fist PC accounts only for shape variation while the second PC account only for pose variation.} 
	\label{PC1_PC2_of_DMO_GPM_no_shape_pose_corr}
\end{figure}

 We built a model using the synthetic training data-set with no shape and pose variation correlation. The model behaved as expected; the first two principal components (PC) accounted for $98.2\%$ of the total variation. The first PC explained shape-only variation without any pose variation and counted for $87.7\%$ and the second explained only pose variation and counted for $10.5\%$ of the total variation. Figure \ref{PC1_PC2_of_DMO_GPM_no_shape_pose_corr} shows the shape variation of the first and second PC from $-3\sigma$ to $+3\sigma$ standard deviations about the mean. It can be observed in the first PC that the pose is fixed at $-12~ rad$ and only the shape parameters $r_1$ and $r_2$ vary from $41.5$ to $77.3 ~mm$ for the first object and from $38.7$ to $4.7 ~mm$ for the second which corresponds to the correlation induced in the training data-set. In the first PC,  the second lollipop head length ($r_2$)  decreases up to its sharp where it starts being unrealistic due to the large sampling range (up to $+3\sigma$)  as indicate by the star (*) in figure \ref{PC1_PC2_of_DMO_GPM_no_shape_pose_corr}. In the second PC the shape parameters are fixed at the mean shape which are $r_1=59.4 ~mm$ for the object 1 and $r_2=21.6 ~mm$ for the second object. The pose parameter $\theta$ varies from $-0.2$ to $-1.5 ~rad$ representing abduction motion in the training data-set. The shape and pose variation explained by two different PCs mirrors the shape and the pose variation across the training data-set confirming that the model represents the training data-set.

\begin{figure}[thpb]
	\centering
	\includegraphics[width=0.48 \textwidth, angle =0 ]{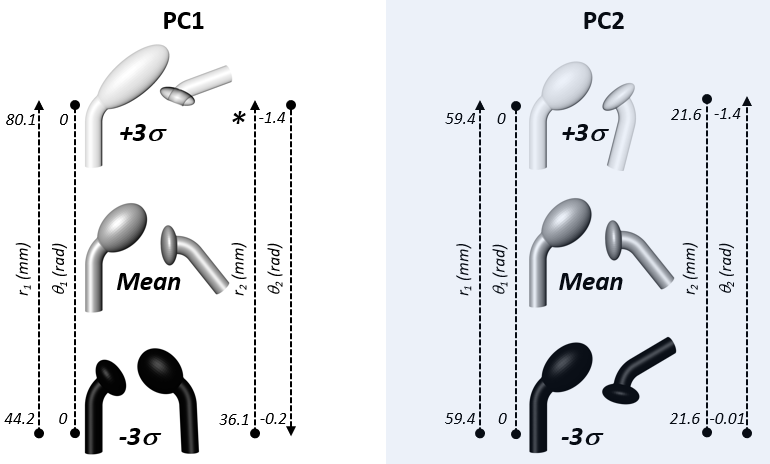}
	\caption{First and second PCs of the DMO-GPM with correlation between shape pose variation. The first PC accounts for both shape and pose variation explaining the correlation between the two set of features. The second PC accounts for pose variation only.} 
	\label{shape_variation for shape-pose model corr}
\end{figure}

We also built a model using synthetic training data in which pose variation was constrained by shape. The model also behaved as expected. The first two PCs accounted for $99\%$ of the total variation. The first PC was shape variation as well as pose variation and accounted for $92.3\%$ and the second explained pose-only variation and accounted for $6.7\%$ of the total variation as shown in Figure \ref{shape_variation for shape-pose model corr}. A visual inspection of the first PC reveals that shape parameters $r_1$ and $r_2$ vary from $44.2$ to $80.1 ~mm$ for the first object and from $36.2$ to $1.6 ~mm$ for the second object and that the pose variation is inversely proportional to the second shape parameter; mirroring the training data-set. In the second PC the shape parameters are fixed at the mean shape parameters which are $59.4 ~mm$ for the first object and $21.6$ for the second object and only the pose parameter $\theta$ varies from $-0.01$ to $-1.4 ~rad$. The shape and pose variation explained by the first PC show a correlation between shape and pose variation similar to that in the training data-set.

\subsubsection{Comparison of DMO-GPM with EDR and SR}
\begin{figure}[thpb]
	\centering
	\includegraphics[width=0.48 \textwidth, angle =0 ]{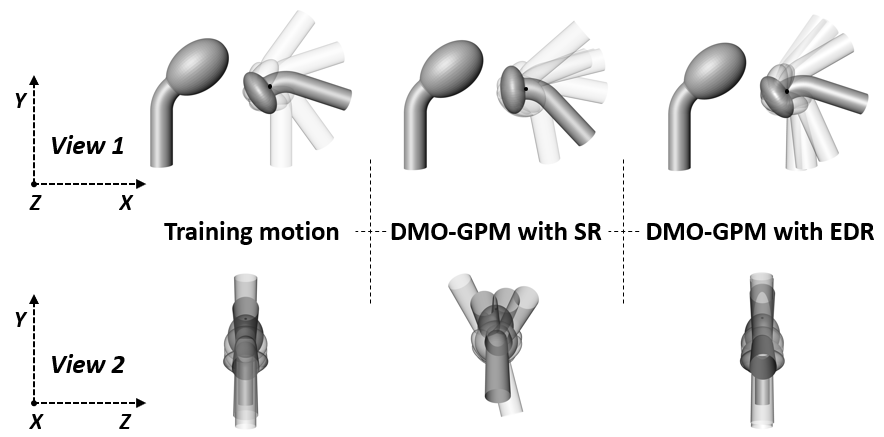}
	\caption{Comparison of the motion sampled from DMO-DPM with EDR with that using DMO-GPR with SR. Left: the motion from the training data. middle: the motion sampled from the model with EDR. Right: the motion sampled from the model with the SR}. 
	\label{DMO-GPM motion vs stardart SSPM motion}
\end{figure}

A DMO-GPM using EDR was compared to the one using the SR to know whether the model with EDR better explains the spatial dynamic of the objects.
We sampled the PC describing the motion simulated and compared this with motion prescribed in the training data. Figure \ref{DMO-GPM motion vs stardart SSPM motion} shows first PC and the third PC which account for motion in the DMO-GPM with EDR and DMO-GPM with the SR, respectively. The motion described by the DMO-GPM is more similar to that prescribed in the training data compared to the motion described by DMO with the SR. For example, the position of the mean shape is at a lower angle in the DMO with SR compared to that in the training motion. The sample poses from $-2$ and $+2$ standard deviations about the mean are completely deviated from the training data motion plane. It can be observed that the mean pose for the model with SR is not different from that of the training data and the pose sampls are not distributed around the mean; this is because the SR does not use object features (is independent of the object features). 

\begin{figure}[thpb]
	\centering
	\includegraphics[width=0.45 \textwidth, angle =0 ]{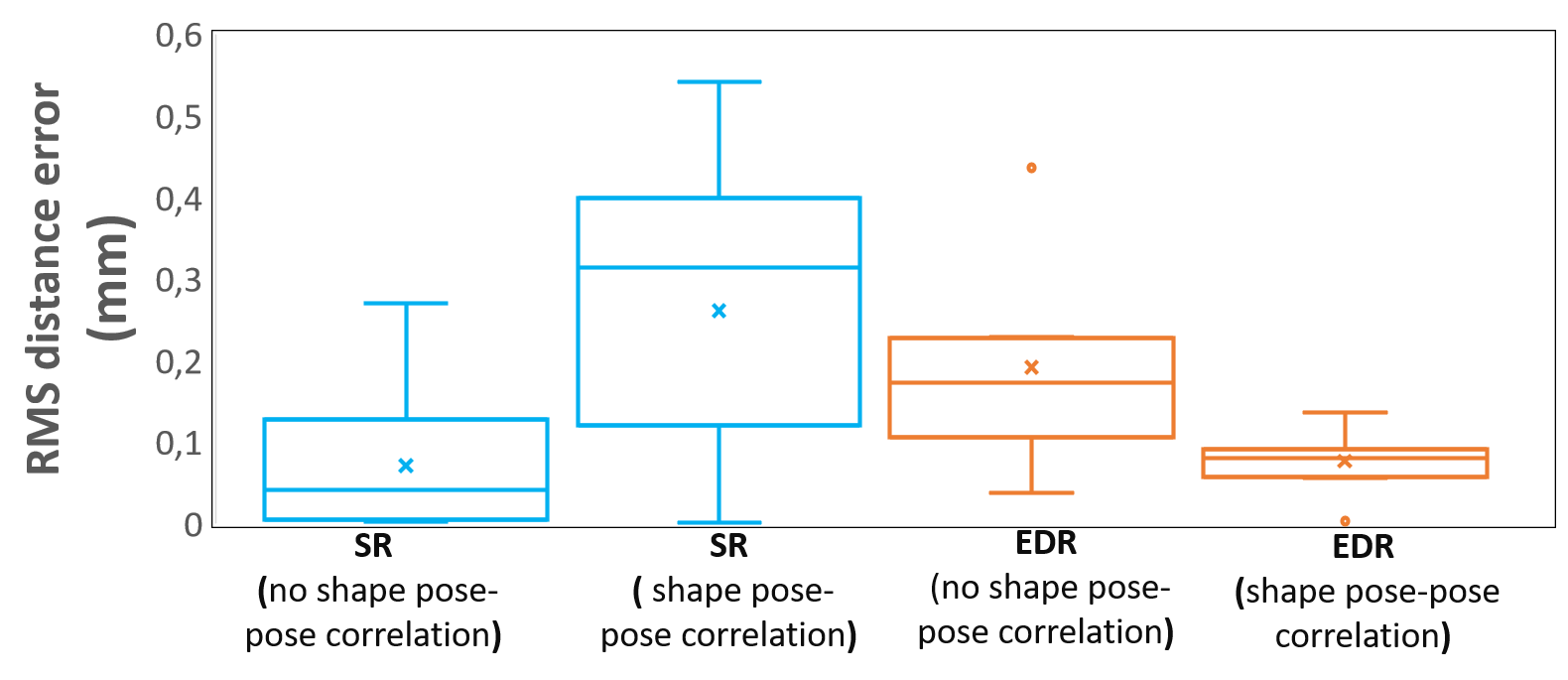}
	\caption{Comparison of the shape plus pose prediction using DMO-GPM with EDR and DMO-GPM with SR. Red:  the RMS errors of the prediction using DMO-GPM with EDR. Blue: the RMS errors of the prediction using DMO-GPM with SR.} 
	\label{DMO-GPM with DEM  vs DMO GPM with standard metric: RMS errors}
\end{figure}

\begin{figure}[thpb]
	\centering
	\includegraphics[width=0.45 \textwidth, angle =0 ]{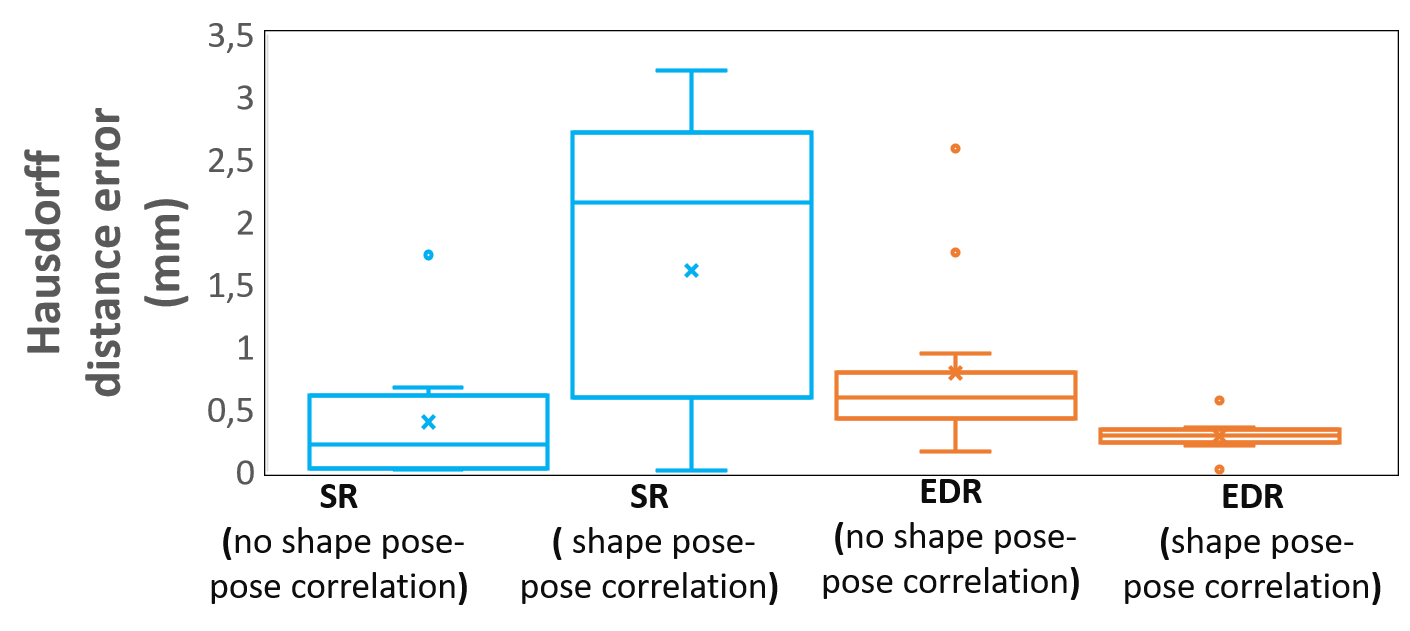}
	\caption{Comparison of the shape plus pose prediction using DMO-GPM with EDR and the one with SR. Red: shows the HD errors of the prediction using DMO-GPM with EDR. Blue: Shows the HD errors of the prediction using DMO-GPM with SR.} 
	\label{MO-GPM with DEM  vs DMO GPM with standard metric: Hausdorff errors}
\end{figure}

We also used the DMO-GPM built with the EDR and the one built with the SR to predict lollipop joints with the second object at various poses, that is, predicting each with type I observation. The same synthetic training examples used to test object correlation above (see table \ref{prediction lillipop using correlation}) were used. The test data-set includes two positions of object 2 relative to the first object: $0 ~rad$ and  $\frac{4}{5}\pi~ rad$ for the models with no shape-pose correlation, and $0~ rad$ and $-\frac{3}{4r_2}\pi~ rad$ for the models with shape-pose correlation. Figures \ref{DMO-GPM with DEM  vs DMO GPM with standard metric: RMS errors} and \ref{MO-GPM with DEM  vs DMO GPM with standard metric: Hausdorff errors} show the box-plots of the surface to surface distance RMS and the HD errors in predicting the second object. The DMO-GPM with EDR outperforms DMO-GPM with SR when there is shape-pose correlation. For no shape-pose correlation, the DMO-GPM with SR performed slightly better than the one with EDR. This slight outperformance occurs even when the motion described by the model is not statistically derived from the training motion, while some poses still belong to the training poses.   

\section{Experiments}

To further demonstrate our framework in clinical and medical image analysis tasks, we applied our method to the shoulder joint; the most complex joint of the human body. The complexity of the shoulder joint can be seen from both the shape and the dynamics points of view. The joint contains the scapula, one of the most complex bones in shape terms; and the humerus, one of the longest bones of the body. This  complexity already confounds the separate morphometric analyses of each bone, let alone when both bones need to be analysed together. Furthermore, in motion terms, the shoulder joint has the most degrees of freedom. This leads to difficulties in segmenting its constituent bones at their relative spatial positions.

\subsection{Prediction of shoulder joint mechanics using DMO-GPM and MCMC}
 We applied DMO-GPMs to shoulder joint data (composed of two articulating bones; scapula and humerus) to demonstrate different scenarios in which the framework could be used. The data consisted of $3D$ mesh surfaces segmented from computed tomography (CT) images of the bilateral shoulders of fresh cadavers collected from the Division of Clinical Anatomy and Biological Anthropology, Faculty of Health Sciences, University of Cape Town, South Africa. Institutional ethics approval was granted for this study (Approval No: HREC 546/2017). A total of $36$ shoulder joints were used; of these data, $8$ were from female and $10$ were from male remains. None presented with shoulder pathology at the time of death. The age of the imaged decendents ranged from 21 to 90 years. The CT volumes were imported into the medical modelling tool (Amira v6.2.0, http://www.fei.com/) for segmentation and 3D surface rendering.

 \subsubsection{Correspondence establishment across shoulder joints}
 In order to establish correspondence across joints within the data-set, we used the surface meshes to create a median virtual shape for the humerus and scapula, separately, using iterative median closest point-Gaussian mixture model (IMCP-GMM) \cite{c29}. The mean-virtual shapes were used as morphologically unbiased templates to built separate GPMMs. These models were used to register all the samples across the data-set using the joint registration method described in section \ref{correspondence}. Table \ref{Shoulder registration errors} shows the average surface to surface RMS and HD between the original meshes and their estimations after the registration process for both humeri and scapulae. The registration errors were low for both humerus and scapula indicating that adequate anatomical correspondence was established across examples in the training data-sets. However, the humerus registration errors were low relative to the scapula registration errors which  could be attributed to the greater shape complexity of the scapula relative to the humerus \cite{c29}. 
 \begin{figure*}[thpb]
	\centering
	\includegraphics[width=1 \textwidth, angle =0 ]{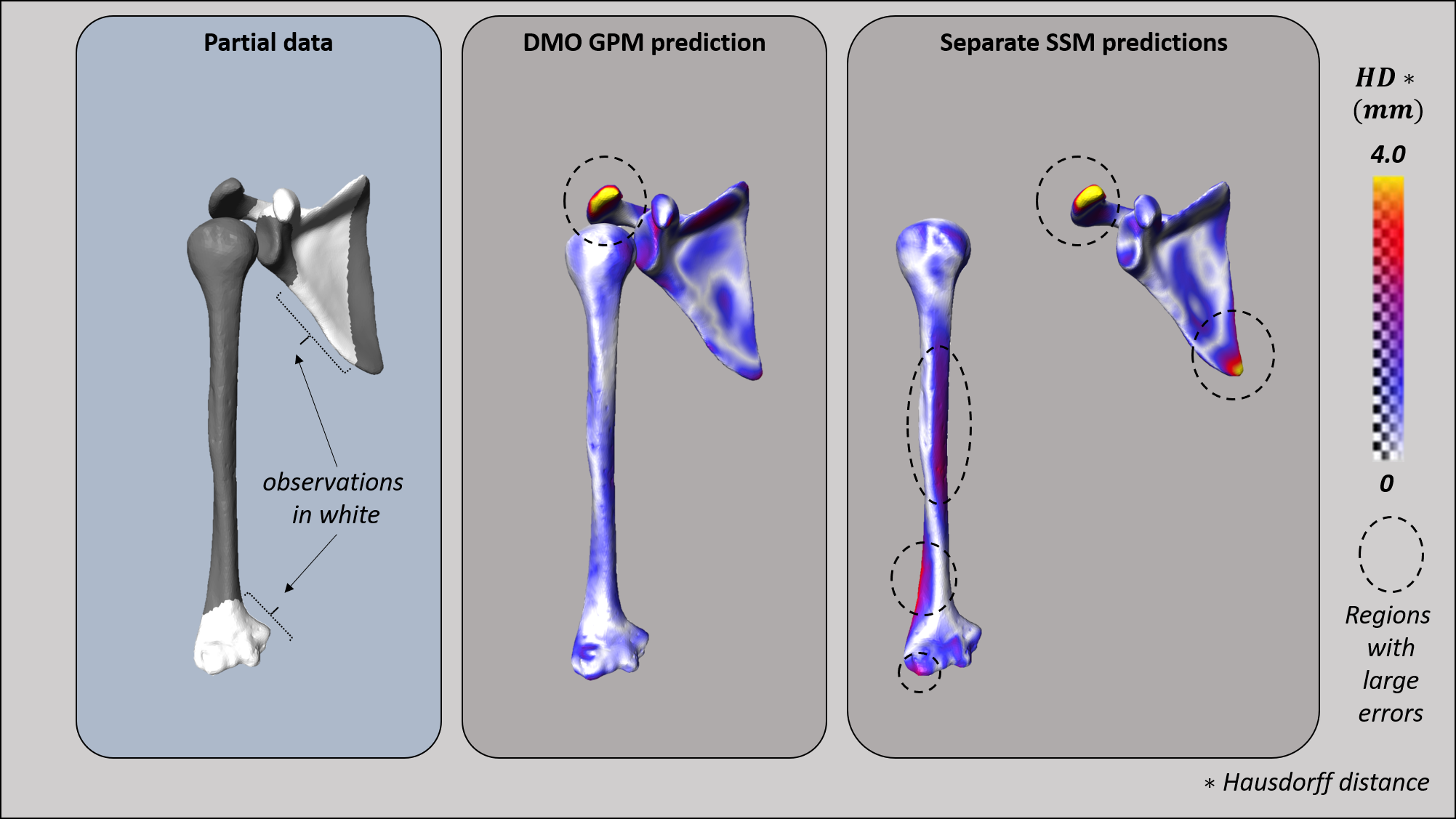}
	\caption{Reconstruction of a partial target scapula and humerus. Observed part of the shoulder  for prediction (white). The Hausdorff distance reconstruction errors show better reconstruction with DMO-GPM (middle) than  with SSMs (right)} 
	\label{best_fit_scapula_humerus}
\end{figure*}
    \begin {table}
\caption {Shoulder joint training data-set registration errors. The average RMS and the HD of scapula and humerus examples.} \label{Shoulder registration errors} 
 \begin{center}
    \begin{tabular}{ |p{2cm}|p{1.5cm}|p{1.5cm}|  }
    \hline
 & RMS (mm) &HD (mm)\\
 \hline
 Humeri  & \textbf{0.73}$\pm$\textbf{0.18} &\textbf{2.62}$\pm$\textbf{0.84}\\
         
  &  & \\
 Scapulae& \textbf{1.11}$\pm$\textbf{0.2}&\textbf{3.01}$\pm$\textbf{0.8}\\
 \hline
\end{tabular}
\end{center}
\end{table}

\subsection{Simulation of shoulder joint motion}
Since the CT scans obtained were from  cadavers, the humeri were in random positions relative to their corresponding scapulae. Some of these positions were not realistic (physiologically), therefore using them for motion analysis would have led to an unrealistic study. To overcome this issue, we simulated artificial motion through a biomechanics protocol.

To simulate the motion, the humerus was considered to be the mobile object while the scapula was fixed. A random shoulder joint from the in-correspondence data-set was used as a template and artificial pose variations were obtained in four steps \cite{c14}: 1) determining the humeral head centre of rotation using a sphere fitting algorithm \cite{c30}; 2) defining the International Society of Biomechanics (ISB) recommended coordinates for both scapula and humerus bones \cite{c9}; 3) obtaining the neutral position of the humerus by aligning the humerus and scapula coordinate systems and then shifting the scapula ISB origin to the glenoid centre; and  4) generating ten poses by moving the humerus around the glenoid centre in the $yz$-plane (simulating abduction). Five poses were generated for each joint $(0, \frac{\pi}{6},\frac{\pi}{6},\frac{\pi}{3},\frac{\pi}{2},\frac{2\pi}{3})$.

\subsection{Predicting shoulder joint shape features}

To demonstrate the performance of our model in predicting shoulder anatomy, we compared the results obtained using DMO-GPM to those using single structure SSMs built as GPMM of the humerus. The prediction was done using MCMC. For this purpose, a marginalised shape-only DMO-GPM of the shoulder was extracted from the global DMO-GPM since only shape features were being predicted. We demonstrated the prediction of the humerus from the observation of the scapula in different scenarios. We created a test data-set, consisting of $10$ shoulder joints ($5$ from the training data and $5$ outside the training set) for shoulder joint shape feature prediction.

\begin{figure}[thpb]
	\centering
	\includegraphics[width=0.45 \textwidth, angle =0 ]{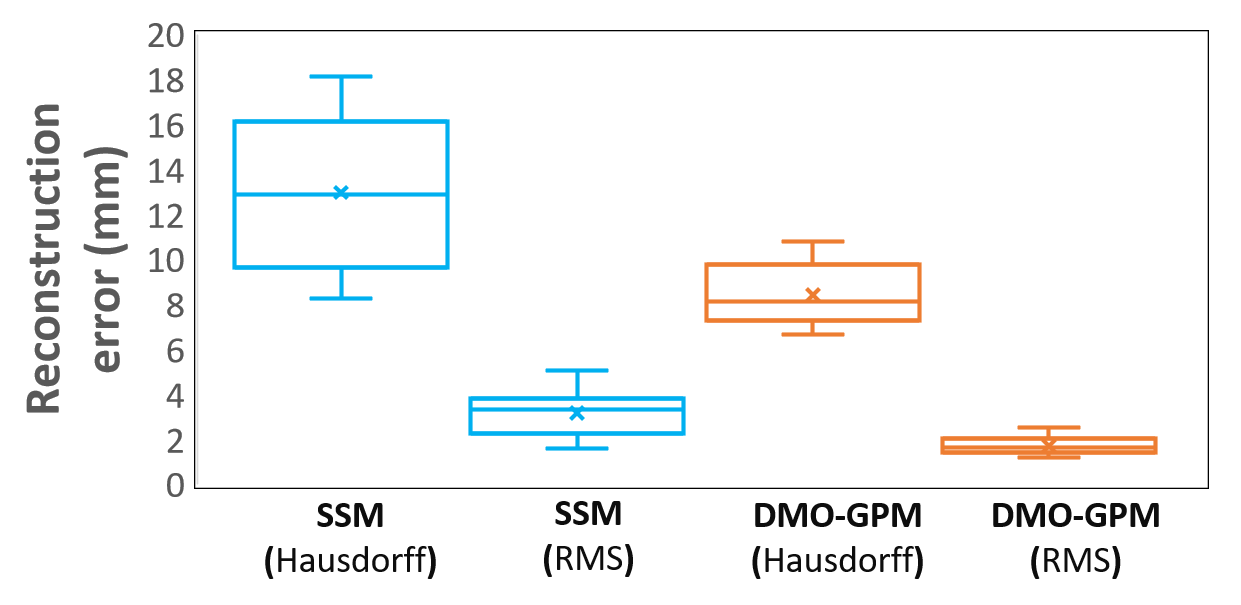}
	\caption{Comparison of the prediction of the humerus with Type III observation SSM with that using DMO-GPM. Blue: the HD (left) and the RMS distance (right) of the prediction using SSM. Red: HD  (left) and the RMS distance (right) of the prediction using DMO-GPM.} 
	\label{fitting_of_humerus_with_two_landmarks_global}
\end{figure}

\begin{figure}[thpb]
	\centering
	\includegraphics[width=0.45 \textwidth, angle =0 ]{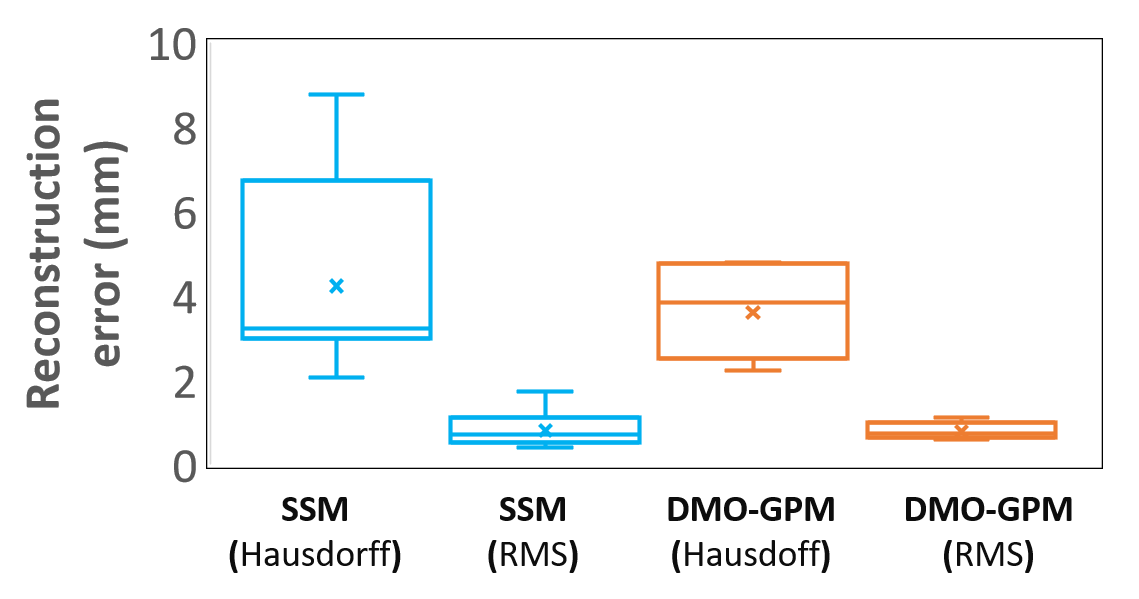}
	\caption{Comparison of the prediction of the humerus with Type II observation SSM with that using DMO-GPM. Blue: the HD  (left) and the RMS distance (right) of the prediction using SSM. Red: the HD  (left) and the RMS distance (right) of the prediction using DMO-GPM} 
	\label{fitting_of_humerus_with_partial_humerus_global}
\end{figure}

First, we predicted the humerus with Type IV observation, that is, only having the scapula of the patient. Table \ref{prediction of humerus without knowledge} shows the errors in predicting the humeri without any information at all using DMO-GPM. Regardless of somewhat large errors, it can be concluded that if there is a strong correlation between the scapula and humerus, a meaningful prediction of one given the other is possible. These predictions result from the correlation that may exist between scapula and humerus shapes captured by our modelling framework. Such predictions would not be possible with single structure models or SSMs. Second, we predicted humeri with Type III observation; two anatomical landmarks (tips of the lateral and medial epicondyle) were selected on the distal humerus targeted for prediction. Our aim here was to predict the humeri when presented with a scapula and the two humerus landmarks using the DMO-GPM. Again we compared the results of the DMO-GPM predictions of the humeri with predictions made using a humerus SSM. Figure \ref{fitting_of_humerus_with_two_landmarks_global} shows the box-plots of prediction errors of the humerus using DMO-GPM and single structure humerus SSM. The prediction errors with DMO-GPM were low compared to errors for predictions with the SSM. The outperformance of the DMO-GPM relative to the single structure SSM could be attributed to the additional humerus knowledge afforded through its corresponding scapula. Finally, humeri were predicted with Type II observation. We simulated humerus fracture scenarios by removing some part of the humeral surfaces and obtained partial humeri. The same testing was performed between DMO-GPM and single structure SSM of the humeri for prediction of the full humeri from partial humeri. Figure \ref{fitting_of_humerus_with_partial_humerus_global} shows the box-plots of the RMS and the HD prediction errors of the humerus using DMO-GPM and SSM. The DMO-GPM outperformed the single structure SSM and figure \ref{best_fit_scapula_humerus} shows a reconstructed scapula with DMO-GPM and individuals SSMs. Again this performance could be due the shape correlation that exists between the humerus and the scapula that can be considered to carry latent information about the humerus.

\begin{figure}[thpb]
	\centering
	\includegraphics[width=0.45 \textwidth, angle =0 ]{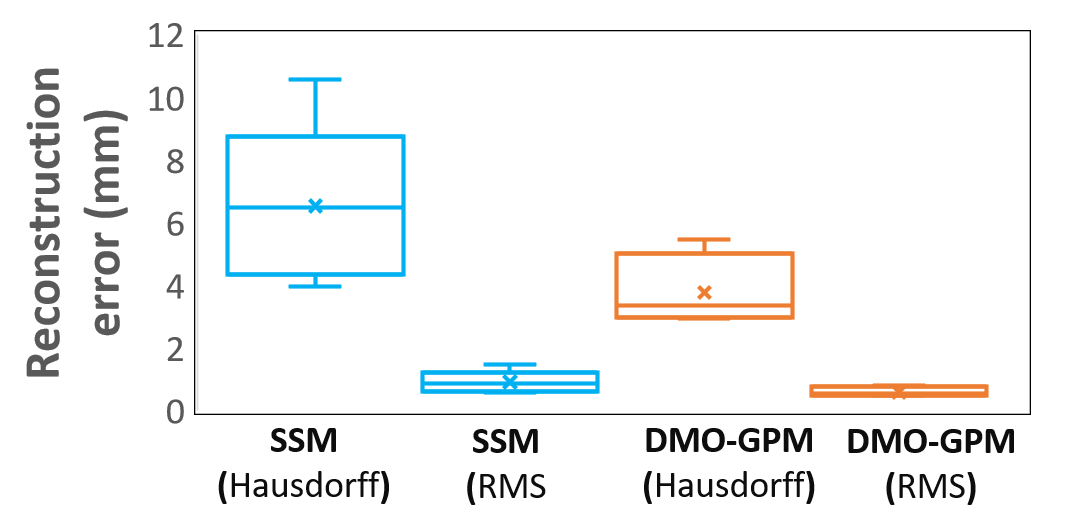}
	\caption{Comparison of the prediction of the scapula with Type II observation SSM with that using DMO-GPM. Blue: the HD (left) and the RMS distance (right) of the prediction using SSM. Red: the HD  (left) and the RMS distance (right) of the prediction using DMO-GPM} 
	\label{fitting_of_scapula_with_partial_scapula_global}
\end{figure}

We predicted scapulae from the observation of partial scapulae using DMO-GPM and GPM; that is, scapula fractures were simulated on the distal part of the scapula. The DMO-GPM and the SSM of the scapula were used to predict those scapulae. Figure \ref{fitting_of_scapula_with_partial_scapula_global} shows the prediction errors, where DMO-GPM outperformed the scapula SSM.

 \begin {table}
\caption {Prediction errors of the humerus with Type IV observations; average RMS and HD} \label{prediction of humerus without knowledge} 
 \begin{center}
\begin{tabular}{|c|c|c|}
\hline
\multicolumn{3}{|c|}{DMO-GPM}\\
\hline
Humerus with Type IV observation & RMS (mm) & HD (mm)\\
\cline{2-3}
 & \textbf{2.9}$\pm$\textbf{1.18}&\textbf{14.0}$\pm$\textbf{5.23}\\ 
\hline
\multicolumn{3}{|c|}{Single structure GPMM}\\
\hline
Humerus with Type IV observation & RMS (mm)& HD (mm)\\
\cline{2-3}
 & \multicolumn{2}{|c|}{Not possible}\\ 
\hline
\end{tabular}
\end{center}
\end{table}

\begin{figure*}[thpb]
	\centering
	\includegraphics[width=1 \textwidth, angle =0 ]{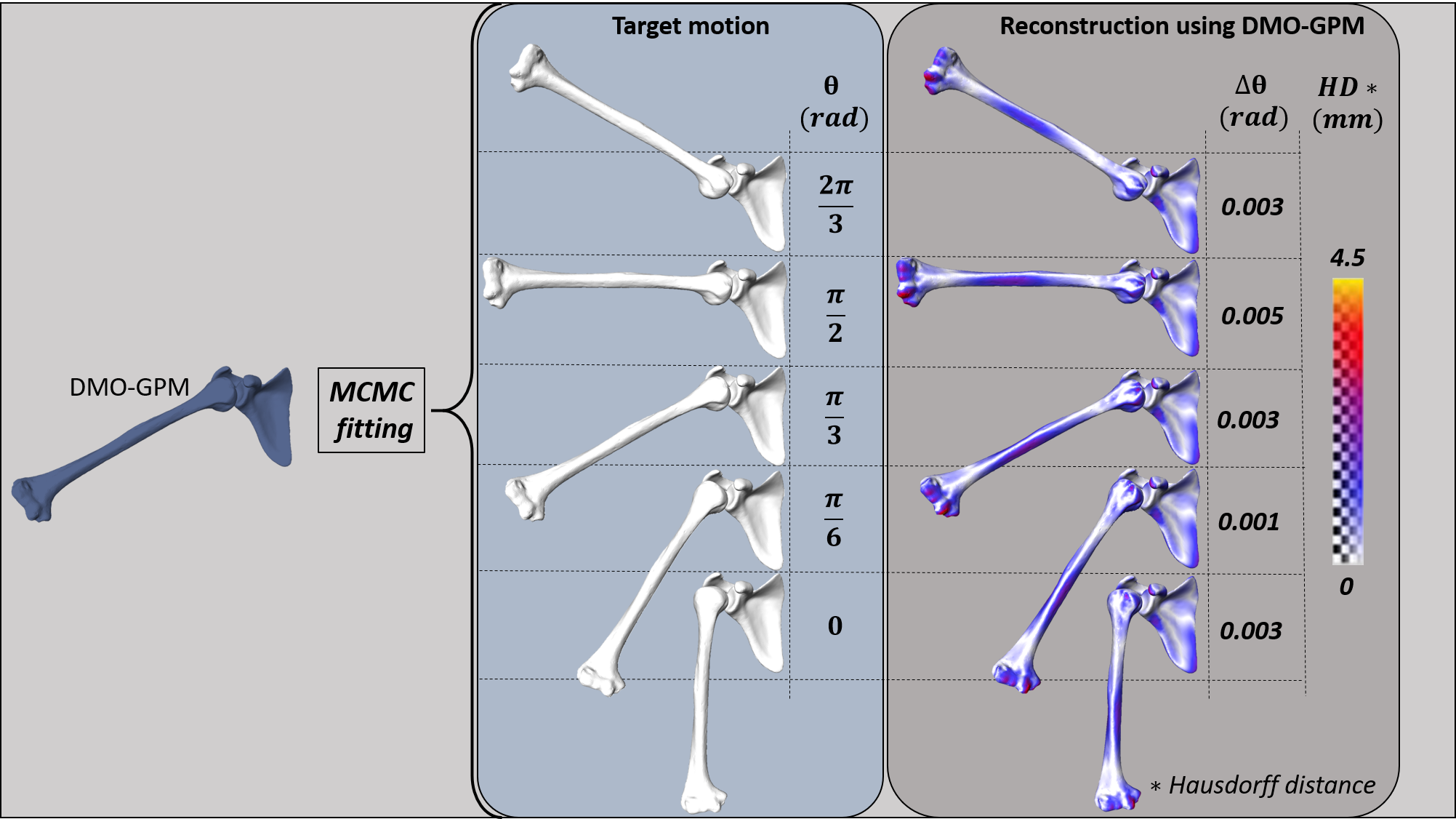}
	\caption{Reconstruction of target shoulders. Left: The DMO-GPM. Middle: target shoulder at various poses (from $0$ to $\dfrac{2\pi}{3} ~rad$) to be predicted. Right: Hausdorff distance  and angle errors for the predicted shoulder using DMO-GPM.} 
	\label{fitmotionDMO}
\end{figure*}

\subsection{Predicting shoulder joint motion}

 \begin{figure}[thpb]
	\centering
	\includegraphics[width=0.45 \textwidth, angle =0 ]{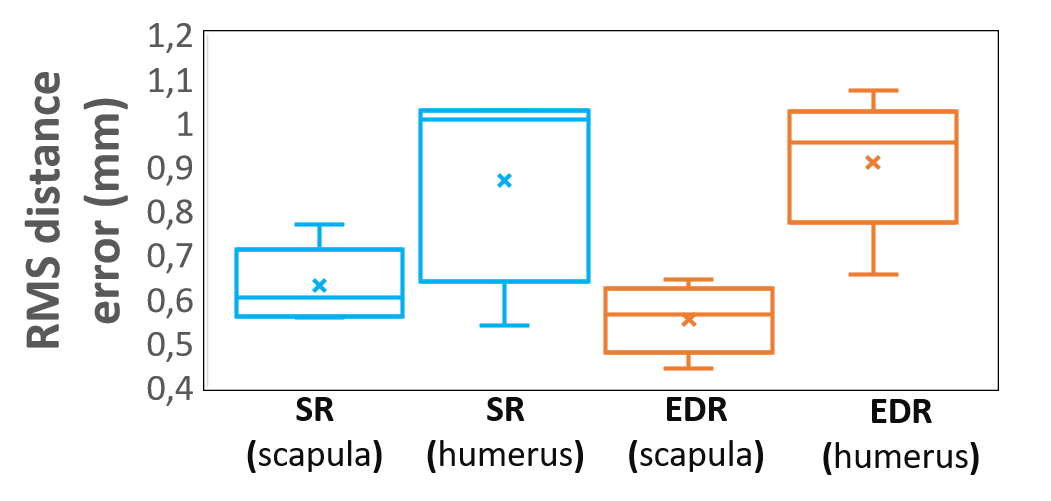}
	\caption{Prediction of the shoulder joint motion.  DMO-GPM with EDR fitting compared to that with SR} 
	\label{shoulder_motion_prediction_rms}
\end{figure}

 To evaluate the performance of the DMO-GPM in predicting training motion, the simulated abduction motion was predicted. The scapula and the humerus (type I observation) were predicted with the humerus at different poses relative to the scapula. Five poses were generated within the motion range $(0,\frac{3}{10}\pi, \frac{1}{2}\pi,\frac{7}{10}\pi,\frac{4}{5}\pi)$, these test poses simulated abduction as in the training data-set.
The prediction was done using DMO-GPM with EDR and with SR in order to compare the performance of the two models in predicting shape at their relative positions. Figure \ref{shoulder_motion_prediction_rms} shows the box-plots of the prediction errors obtained using pose marginalised DMO-GPM with EDR and with SR. The DMO-GPM with the EDR outperformed that with SR. For both scapula and humerus, the median error of the prediction with the EDR model was less than that with SR. Figure \ref{fitmotionDMO} shows the reconstructed shoulder using the DMO-GPM.

\section{Conclusion}



Understanding patterns and explaining them in a statistical framework so that we can induce meaningful observations is a vast area of research. This study presents a framework that tackles the existing problems in a combined statistical representation of multi-object shape and pose structures by providing a homogeneous yet simple way of building data-driven generative models. The common aim in statistical modelling of multi-objects is to efficiently explain shape variability and spatial dynamics across a population of multi-objects in a low dimensional space. This study focused bringing the two sets of feature classes involved (shape and pose) into a common linearized space, thus finding the homogeneous analysis space. While multiple methods exist to statistically represent the shape variations in linear space, its pose counterpart is often ignored in the literature, which may lead to incorrect variability of such shape-pose models. The proposed framework not only uses a homogeneous and linear statistical representation of both shape and pose features, it also embeds the mode in a continuous domain thereby making the model generative. This type of framework in especially needed in the medical imaging domain where structures of interest can be human articulating joints. Human joint structures represent non-linear and non-homogeneous feature classes in terms of its bone shapes and poses. Optimal and homogeneous statistical representation of the variation across these structures using a data-driven modelling framework is of utmost importance to maintain the clinical accuracy of the inferences drawn from the model. In this study, we have illustrated that DMO-GPM modeling framework can have valid representations of shape and pose combined for multi-object structures. 


 We have provided a systematic validation of the DMO-GPM framework using closely-controlled synthetic data (lollipop) and illustrated its application is experimental medical data (shoulder joint). This framework is able to incorporate more than two objects in a compact representation using Gaussian process which means the developed models provide a continuous analysis space with variations  of shape and spatial dynamics. Marginalizing DMO-GPMs allow us to analyze and/or predict shape-only or pose-only or combined extractions/generations. This important characteristic of DMO-GPMs allow us to implement the framework in multiple applications ranging from pre-surgical planning to population based implant designs.

Results of validation with a synthetic data-set provide the efficiency of the DMO-GPM framework in explaining correlations across a given population of dynamic multi-objects. The DMO-GPM framework can accurately explain shape correlation between objects and correlation between shape and spatial variation, an advantage of the multi-object models in prediction compared to single structure SSMs. Furthermore, DMO-GPM also preserves the optimal representation of 3D motion using the EDR approach as illustrated on the non-compact lollipop data-set. We also applied our framework on human shoulder data to address one of the challenges in medical image processing, that is, prediction  from  observations. DMO-GPM illustrated the prediction ability of one of the shoulder bones from the observation of the other. The prediction would be more accurate if more observations - partial or full - are available. For example, estimating the humerus from the observation of the clavicle and the scapula would be better than from the observation of the scapula only. This could be of interest in surgical settings such as shoulder arthroplasty, especially  when one of the joint bones is partially or completely missing (due to trauma).    

The challenge of automatic assessment of pose variation in medical image analysis can be due to the inability to standardize pose during image acquisition. This is true for the same individual with multiple images or different individuals. We evaluated our model in this scenario by prediction of a shoulder joint with the humerus at different relative positions to scapula, around a predetermined joint center, simulating abduction. Results suggest that DMO-GPM can  be incorporated into such scenarios for automatic segmentation pipelines. Future work would involve extension of DMO-GPM with an intensity model \cite{c23}. This would allow the use of the model for automatic segmentation of $3D$ images (MRI and CT scans), especially dynamic MRI, with due consideration to pose variations. Furthermore, DMO-GPM could be used to optimize surgery. For example, by exploiting anatomical correlation between joint objects,  accurate joint segmentation could be achieved for the design of the better joint implants. Additionally, parametric optimization could be investigated, which is computationally less expensive than MCMC and could improve the clinical utility of the models. Moreover, the proposed framework could be used to answer biological questions.

\appendices
\section{Notations}
\begin{tabular}{ll}
$k_{\Omega}^{(s,\sigma)}$ & Gaussian kernel defined in $\Omega$ with\\
& the scale $s$ and variance $\sigma$
\end{tabular}

\begin{tabular}{ll}
$\mathcal{N}(0,1)$ & normal distribution with\\
&zero and unit covariance\\
$S^j_i$ & $j^{th}$ object of the $i^{th}$ example \\
&in the training data \\
$n$  & number of the training examples\\
$N$ & number of objects constituting a\\
&training example\\
 $h^j$ & rigid transformation assisted to the \\
 $j^th$ &shape sample from the DMO-GPM\\ 
$SE(3) $ & space of similarity transformations\\
$X^i$ & concatenated vector of shape and pose\\ 
&features of the $i^{th}$ example\\
$e^k$ & the combined eigenvector of shape ($e^k_S$) \\
& and pose ($e^k_p$) for SSPMs\\
$\mu$ & population mean \\
$\Sigma$ & population covarinace \\
$\alpha $ & Gaussian random vector\\
$X_{\alpha_k}$ & shape sample from SSPMs\\
$h_{\alpha_k}$ & pose sample from SSPMs\\
$u_i$ & deformation field associated to\\
&the $i^{th}$ example\\
$\Omega_S$ & reference shape domain\\
$\Omega_K$ & reference pose domain\\
$\Gamma_S$& reference shape\\
$\lambda_m$ & $m^{th}$ eigenvalue of the GPMMs\\
$\phi_m$ & $m^{th}$ eigenvector of the GPMMs\\
$S^j_0$ & $j^{th}$ object of the fixed object family\\ $\Gamma^{j_0}$ & reference of the fixed object family\\
$u^{MO}$ & multi object deformation field\\
$S^j$ & $j^{th}$ object of the training example\\
$T$ & rigid transformation between  \\
        &the reference $\Gamma_S^j$ and the object $S^j$\\
 $T^j$ & rigid transformation between  \\
        &the reference $\Gamma^j_0$ and the object $S^{j_0}$\\
$S^j_r$ & registered object corresponding \\
&to the object $S^j$\\
$X$ &  multi-object concatenated vector of shape \\
&and spatial position\\
$K^i$ & spatial position displacement field of the\\
&$i^{th}$ example \\
$X_K$ & spatial position component of $X$\\
$X_S$ & shape component of $X$\\
$\sigma^j_i$ & energy displacement representation\\
&associated to $S_i^j$\\
$\Omega_K^j$ & reference associated to the $K^j$\\
$u_i^{MO}$ & multi-object deformation field\\
& associated to $i^{th}$ example \\
$\mu_{MO}$ & DMO-GPM mean\\
$\phi^S_m$ & shape component of $\Psi^{MO}_m$\\
$\phi^K_m$ & pose component of $\Psi^{MO}_m$\\
$T_o$& target multi-object example to be predicted\\
$r_1$ &  shape parameter of the $1^{st}$ synthetic object\\
$r_2$ &  shape parameter of the $2^{nd}$ synthetic object\\
$\varphi$ & Euler angles in X-axis\\
$\theta$ & Euler angle in  Y-axis\\
 $\psi$ & Euler angle in  Z-axis\\
$p_{\textrm{MO}}$  &  DMO-GPM\\
$\Gamma_S^j$ & reference of the $j$'s objects \\
$\vec{\Gamma}_S^j$ & a vector coordinate of the points of $\Gamma_S^j$ \\
$h^j_i$ & rigid transformation between  \\
 &the reference $\Gamma_S^j$ and the object $S^j_i$
\end{tabular}

\ifCLASSOPTIONcompsoc
  \section*{Acknowledgments}
\else
  \section*{Acknowledgment}
\fi

The authors would like to thank Marcel L\"uthi (Department of Mathematics 
and Computer Science, University of Basel, Basel, Switzerland) for invaluable input and discussion in preparation of this study.

\ifCLASSOPTIONcaptionsoff
  \newpage
\fi



%

\end{document}